\documentclass[9pt]{article}
\usepackage{spconf,amsmath,graphicx}

\usepackage{amssymb}
\usepackage{graphicx,psfrag} 
\usepackage{epsfig} 
\usepackage{esdiff} 
\usepackage{amsmath} 
\usepackage{xfrac}   
\usepackage{subfigure}  
\usepackage{pst-sigsys} 
\usepackage{algorithm}
\usepackage{cite}
\usepackage{algorithmic}
\usepackage{colortbl}
\usepackage{setspace}
\usepackage{color}

\newtheorem{lem}{Proposition}

\newtheorem{defn}{Definition}

\newcommand{\beq}{\begin{equation}}
\newcommand{\eeq}{\end{equation}}
\newcommand{\beqa}{\begin{eqnarray}}
\newcommand{\eeqa}{\end{eqnarray}}
\newcommand{\ben}{\begin{enumerate}}
\newcommand{\een}{\end{enumerate}}

\title{A SPLIT-AND-MERGE DICTIONARY LEARNING ALGORITHM FOR SPARSE REPRESENTATION}
\name{Subhadip Mukherjee and Chandra Sekhar Seelamantula}
\address{Department of Electrical Engineering, Indian Institute of Science, Bangalore 560012, India\\Emails: subhadip@ee.iisc.ernet.in, chandra.sekhar@ieee.org}
\begin{document}
%
\maketitle
\small
\begin{abstract}
In big data image/video analytics, we encounter the problem of learning an overcomplete dictionary for sparse representation from a large training dataset, which can not be processed at once because of storage and computational constraints.~To tackle the problem of dictionary learning in such scenarios, we propose an algorithm for  parallel dictionary learning. The fundamental idea behind the algorithm is to learn a sparse representation in two phases.~In the first phase, the whole training dataset is partitioned into small non-overlapping subsets, and a dictionary is trained independently on each small database. In the second phase, the dictionaries are merged to form a global dictionary. We show that the proposed algorithm is efficient in its usage of memory and computational complexity, and performs on par with the standard learning strategy operating on the entire data at a time. As an application, we consider the problem of image denoising. We present a comparative analysis of our algorithm with the standard learning techniques, that use the entire database at a time, in terms of training and denoising performance. We observe that the split-and-merge algorithm results in a remarkable reduction of training time, without significantly affecting the denoising performance. 
\end{abstract}

%
\section{Introduction}
In recent years, the problem of learning signal-dependent dictionaries for sparse representation has gained attention in the sparse signal processing research community.  The principal idea behind the problem is to learn a dictionary from a pool of training signals/images that are most likely to occur in a particular application. In many image processing applications, we encounter the issue of training a dictionary over a  dataset of large size. The computational as well as storage burden to handle such big datasets at a time using the standard learning strategy is unacceptably high, and hence calls for parallel processing. In the standard technique, the entire dataset is used at a time and the dictionary is trained by means of an alternating minimization strategy. Each iteration of the standard technique comprises two phases, namely, sparse coding and dictionary update. In the sparse coding phase, one updates the sparse representation for a fixed dictionary, and in the next step, the dictionary is updated for a fixed sparse coefficient matrix. The dictionary update is performed using the classical least squares based Method of Optimal Directions (MOD) \cite{engan}, whereas the sparse coding is accomplished by employing the orthogonal matching pursuit (OMP)  \cite{tropp_omp} algorithm. In this paper, we propose a methodology for learning the dictionary using a parallel approach, which exploits the fact that dictionary learning for the purpose of sparse representation can be done in multiple stages. We refer to the algorithm as the split-and-merge algorithm. Given a big training dataset, we split it into small non-intersecting subsets, and train a dictionary over all the smaller datasets. We refer to the dictionaries trained over smaller datasets as the local dictionaries. The local dictionaries contain equal or lesser number of atoms than what we intend to have in the global dictionary. Finally, we merge the local dictionaries to construct a single global one that represents the entire dataset sparsely. Merging is accomplished by solving another dictionary learning problem, where we search for a dictionary that can represent the atoms in the local dictionaries using a sparse representation. Our analysis shows that the resulting global dictionary represents the whole dataset with a sparse coefficient matrix. To develop the basic idea behind the approach, we define a sparse model for a signal as follows: 
\label{sec:intro}
\begin{defn}
A signal $y \in \mathbb{R}^m$ is said to follow a sparse model $\mathcal{S}\left( D,x,\epsilon,s \right)$ if there exist an overcomplete dictionary $D \in \mathbb{R}^{m \times K}$ and a vector $x \in \mathbb{R}^K$ with $\left\| x \right\|_0 \leq s$ such that $\left \| y-Dx \right \|_2 \leq \epsilon$, for some $\epsilon >0$. We denote it as $y \in \mathcal{S}\left( D,x,\epsilon,s \right)$.
\end{defn}
Equipped with this definition, we state the following proposition, which forms the central idea behind the proposed algorithm.
\begin{lem}
Let $y \in \mathcal{S}\left( \tilde{D},x_1,\epsilon_1,s_1 \right)$ and each column $\tilde{d}_j$ of $\tilde{D}$ be in $\mathcal{S}\left( D,z_j,\frac{\epsilon_2}{\left\| x_1 \right\|_2},s_2 \right)$. Then $y \in \mathcal{S}\left( D,Zx_1,K\epsilon_2+\epsilon_1,s_1 s_2 \right)$, where $Z$ is a matrix constructed by stacking $z_j$s on the columns. 
\end{lem}
 The proof of the proposition is given in Appendix A. This proposition suggests that the problem of dictionary learning can be solved in two stages, provided that the sparsity levels are appropriately chosen at each stage. Before we present our algorithm formally, we briefly review some recent literature.

\subsection{Review of some recent literature on dictionary learning}
Initial contributions to the solution of dictionary learning problem were made by Aharon et al. \cite{elad1}.~They proposed an algorithm, namely, the K-SVD, in which one updates the dictionary atoms in a sequential manner, using the singular vectors of the error matrix resulting from the absence of that particular atom. Aharon et al. deployed this algorithm for the task of image denoising in \cite{elad3}. Yang et al. \cite{yang} used the idea of dictionary learning for image super-resolution.~Abolghasemi et al. \cite{vahid} proposed an adaptive dictionary learning method for blind image source separation.~A greedy adaptive dictionary learning algorithm was developed in \cite{jafari} to find sparse atoms for speech signals.~Dai et al. \cite{dai,dai1} addressed the problem of slow convergence of the training algorithms because of singular points in the dictionary update stage and proposed a simultaneous codeword optimization (SimCO) formulation to alleviate the problem due to singularity. This formulation offers a generalization over the least-squares based MOD algorithm and the K-SVD algorithm, that is, both algorithms become special cases of the SimCO formulation. The problem of learning structured dictionaries was addressed in \cite{ramirez,barchiesi1,barchiesi,mailhe}. Recently, the problem of distributed dictionary learning over sensor networks has been addressed by Chainais et al. \cite{chainais}, who proposed a distributed block coordinate descent approach.~Their solution can be adapted to various matrix factorization problems.

%

\section{Problem Formulation and Proposed Algorithm}
\label{sec_prob_form}
Given a set of $N$ training vectors $\left\{ y_i \right\}_{i=1}^{N} \in \mathbb{R}^{m}$, where $N$ is large, our main objective is to learn a dictionary $D\in \mathbb{R}^{m \times K}$, $m<K$, such that $D$ represents each $y_i$ with an $s$-sparse coefficient vector $x_i$, that is, $\left\| y_i -Dx_i \right\|_2 \leq \epsilon $ with $x_i$ satisfying $\left\| x_i \right\|_0 \leq s \ll K, \forall i$. Let  $Y \in \mathbb{R}^{m \times N}$ denote the matrix formed by stacking the training vectors $y_i$ on the columns. 

\noindent We propose a parallel learning approach, referred to as the split-and-merge, to solve this problem. First, we partition the training dataset $Y$ randomly into $L$ smaller disjoint datasets, each of size $n=\frac{N}{L}$, and train a dictionary on each small dataset. Let the dictionary trained on dataset $t$, $1 \leq t \leq L$, be denoted by $D^{(t)}\in \mathbb{R}^{m \times K_1}$. In order to obtain a global dictionary from the local dictionaries, we form a new dataset by stacking the dictionaries $D^{(t)}$ on the columns (with proper scaling), and train a dictionary over this new dataset. Our analysis shows that this final dictionary represents the entire dataset with desired sparsity level, provided that the sparsity is chosen appropriately for the subproblems. The size $K_1$ of the local dictionaries is usually chosen such that $K_1\leq K$, and $K_1L$ is approximately equal to $\frac{N}{L}$, to ensure that the computational overhead of the merging step is of the same order as each of the smaller dictionary learning subproblems. We describe the procedure formally in Algorithm 1. The sparsity levels $s_1$ and $s_2$ in steps 2 and 3 of the algorithm are so chosen that $s=s_1s_2$, where $s$ is the desired sparsity level for the overall dataset. By invoking Proposition 1, we observe that every training signal in the overall dataset can be represented with  an $s$-sparse representation using the global dictionary. A comparison of the computational complexity with the standard training approach is carried out in Appendix B. 

%

\begin{algorithm}[t]
\caption{\small Split-and-merge algorithm to learn a dictionary $D \in \mathbb{R}^{m \times K}$ from a database $Y \in \mathbb{R}^{m \times N}$, such that $D$ represents each column of the data matrix $Y$ using an $s$-sparse coefficient vector.}
\begin{algorithmic}
\STATE {\bf 1. Split the training dataset}: Decompose $Y$ randomly into $L$ non-overlapping datasets $\left \{Y^{(t)} \right \}_{t=1}^{L}$, each of size $m \times n$, where $n=\frac{N}{L}$. 

\STATE{\bf 2. Train a dictionary on each dataset}: Learn a dictionary $D^{(t)} \in \mathbb{R}^{m \times K_1}$, $K_1 \leq K$, to represent the columns of $Y^{(t)}$ with a sparsity level of $s_1<s$.

\STATE {\bf 3. Merge the dictionaries into a single one}: Construct a new data set $\tilde{Y} = \left[  \tilde{D}^{(1)}  \tilde{D}^{(2)}  \cdots   \tilde{D}^{(L)}  \right]\in \mathbb{R}^{m \times K_1L}$, where $\tilde{D}^{(t)}=\left\| X^{(t)} \right\|_FD^{(t)}$. Learn the dictionary $D \in \mathbb{R}^{m \times K}$, which represents the columns of $\tilde{Y}$ with sparsity $s_2=\frac{s}{s_1}$.

\end{algorithmic}
\end{algorithm}

\section{EXPERIMENTAL RESULTS}
\label{sec_simulation}
We present the results of the experiments performed on synthesized signals as well as real images. 

\begin{figure*}[t]
\centering
\subfigure[Ground truth clean image]{
\includegraphics[width=3.7cm]{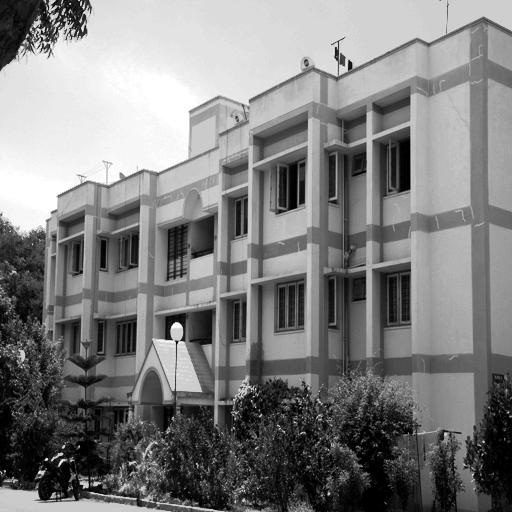}
}
\subfigure[Noisy input image, \text{\,\,\,\,\,\,\,\,\,\,\,\,\,\,\,} PSNR $20.17$ dB]{
\includegraphics[width=3.7cm]{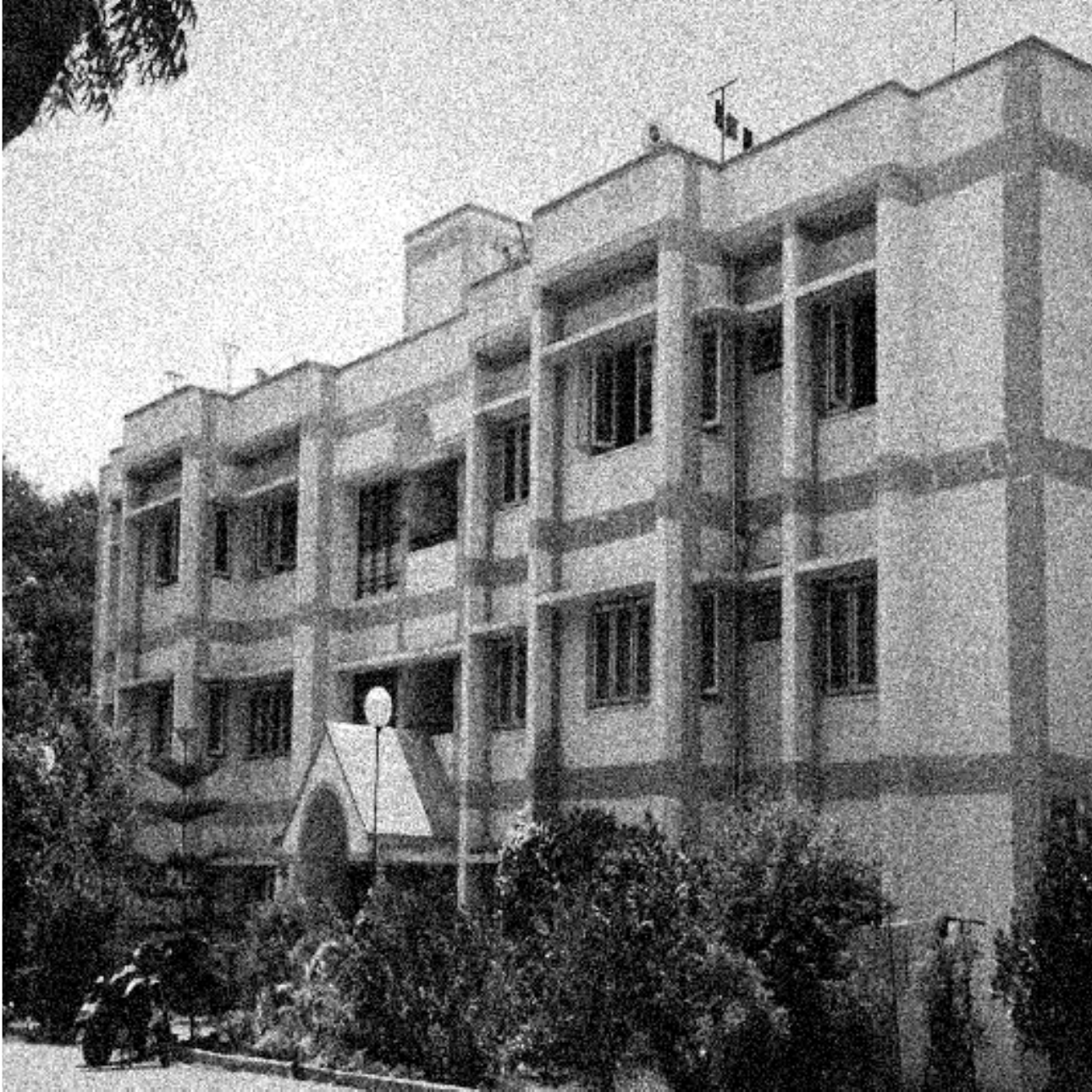}
}
\subfigure[Image denoised using the conventionally trained dictionary, PSNR $28.22$ dB]{
\includegraphics[width=3.7cm]{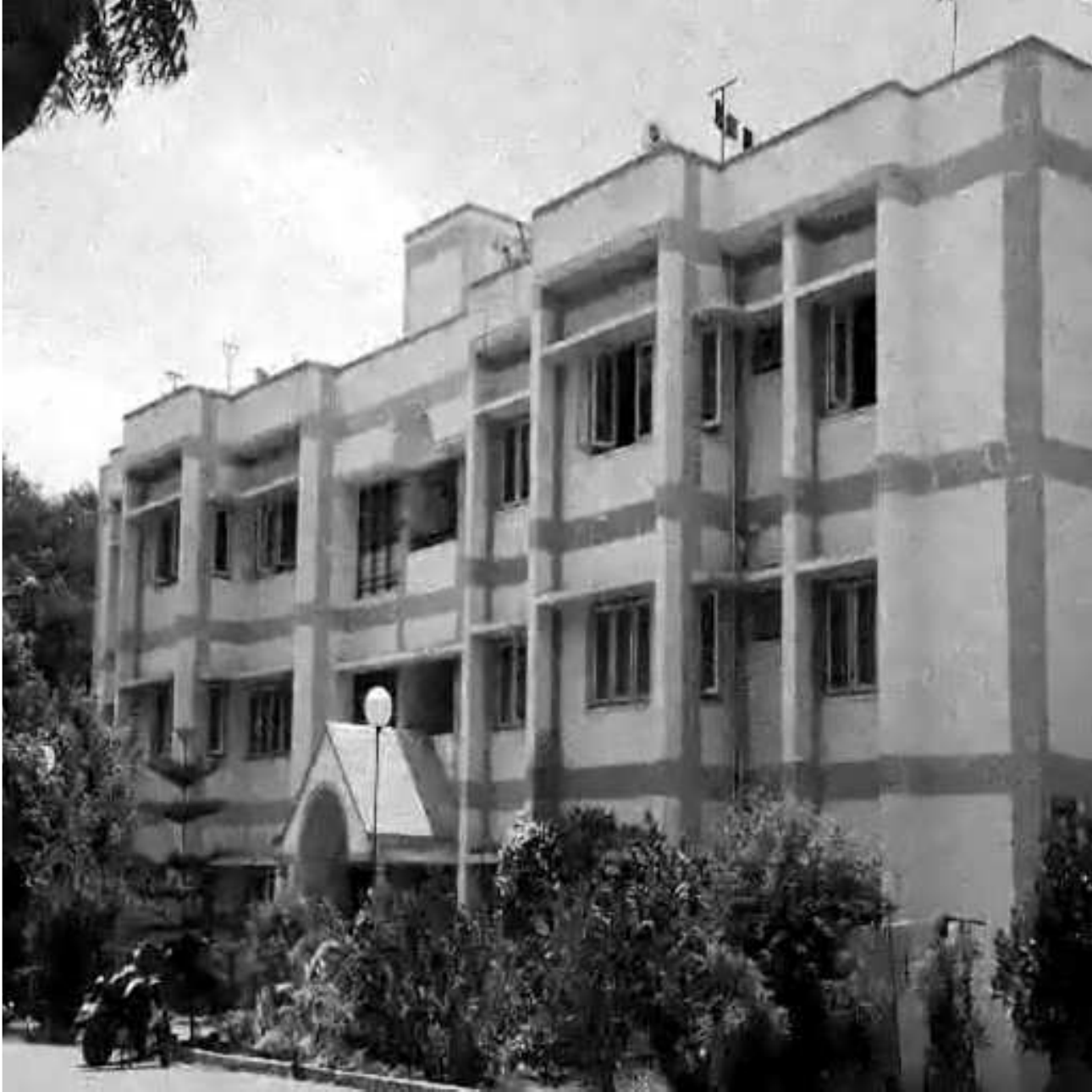}
}
\subfigure[Image denoised using the dictionary trained with Algorithm 1, PSNR $28.07$ dB]{
\includegraphics[width=3.7cm]{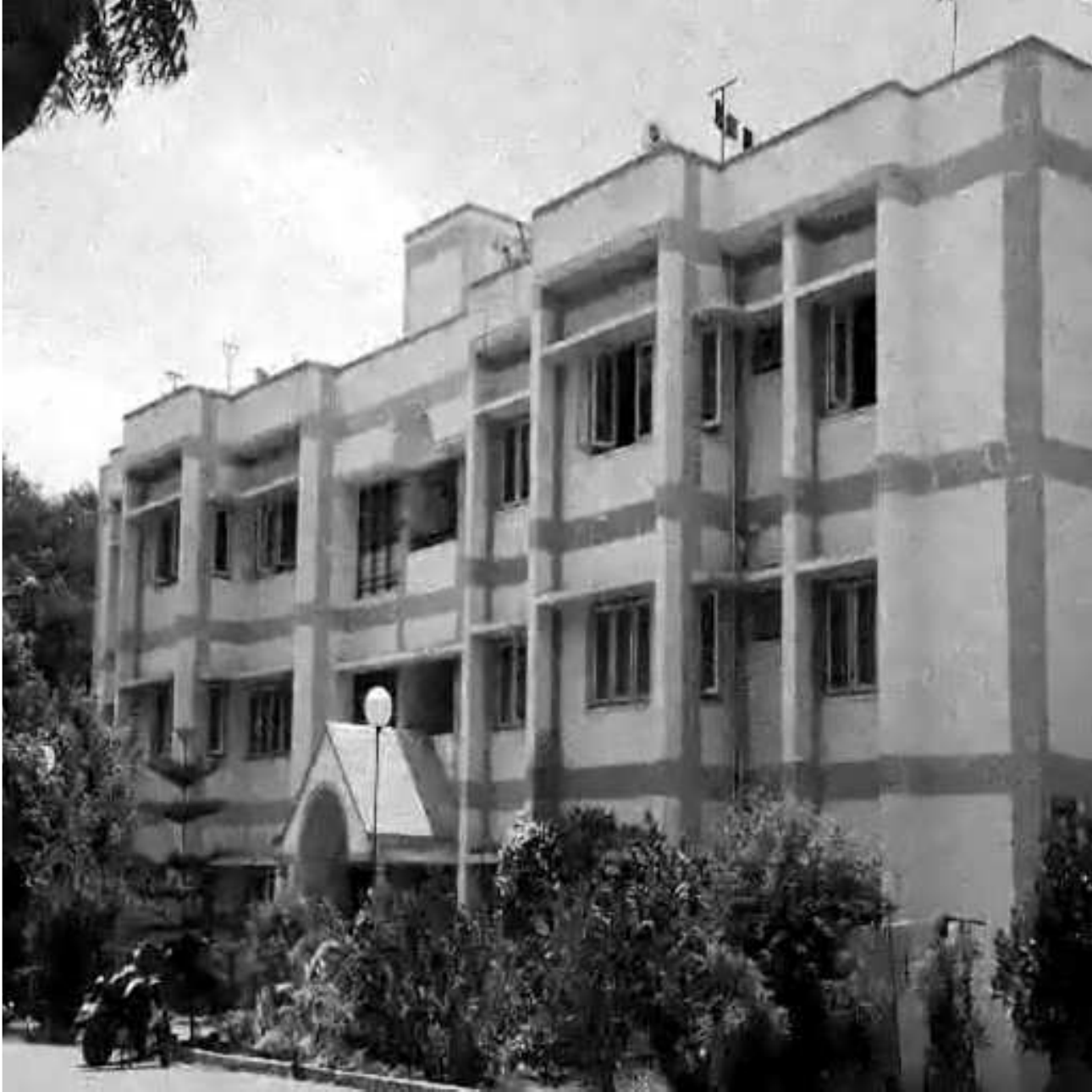}
}\\
\subfigure[Ground truth clean image]{
\includegraphics[width=3.7cm]{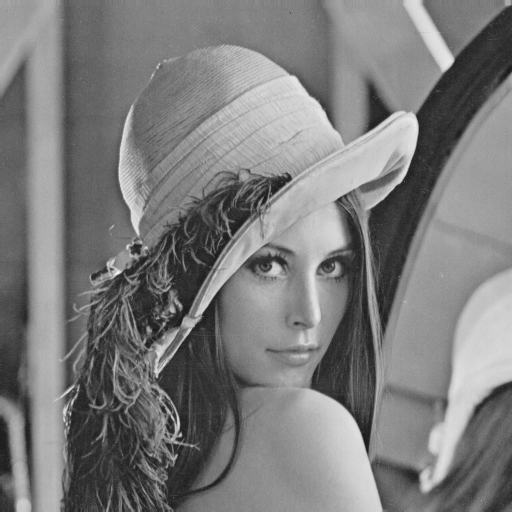}
}
\subfigure[Noisy~input~image, \text{\,\,\,\,\,\,\,\,\,\,\,\,\,\,\,} PSNR $20.17$ dB]{
\includegraphics[width=3.7cm]{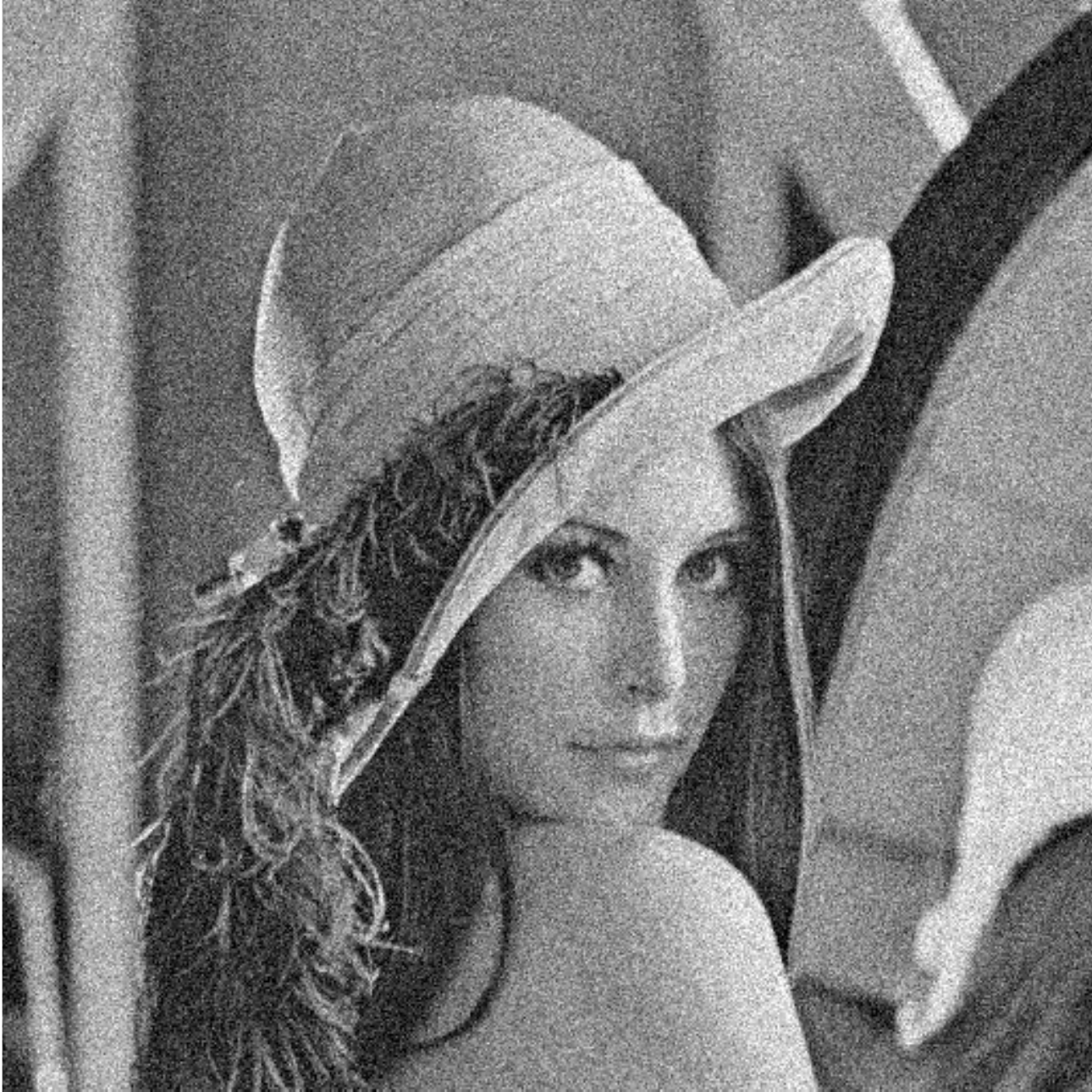}
}
\subfigure[Image denoised using the conventionally trained dictionary, PSNR $32.21$ dB]{
\includegraphics[width=3.7cm]{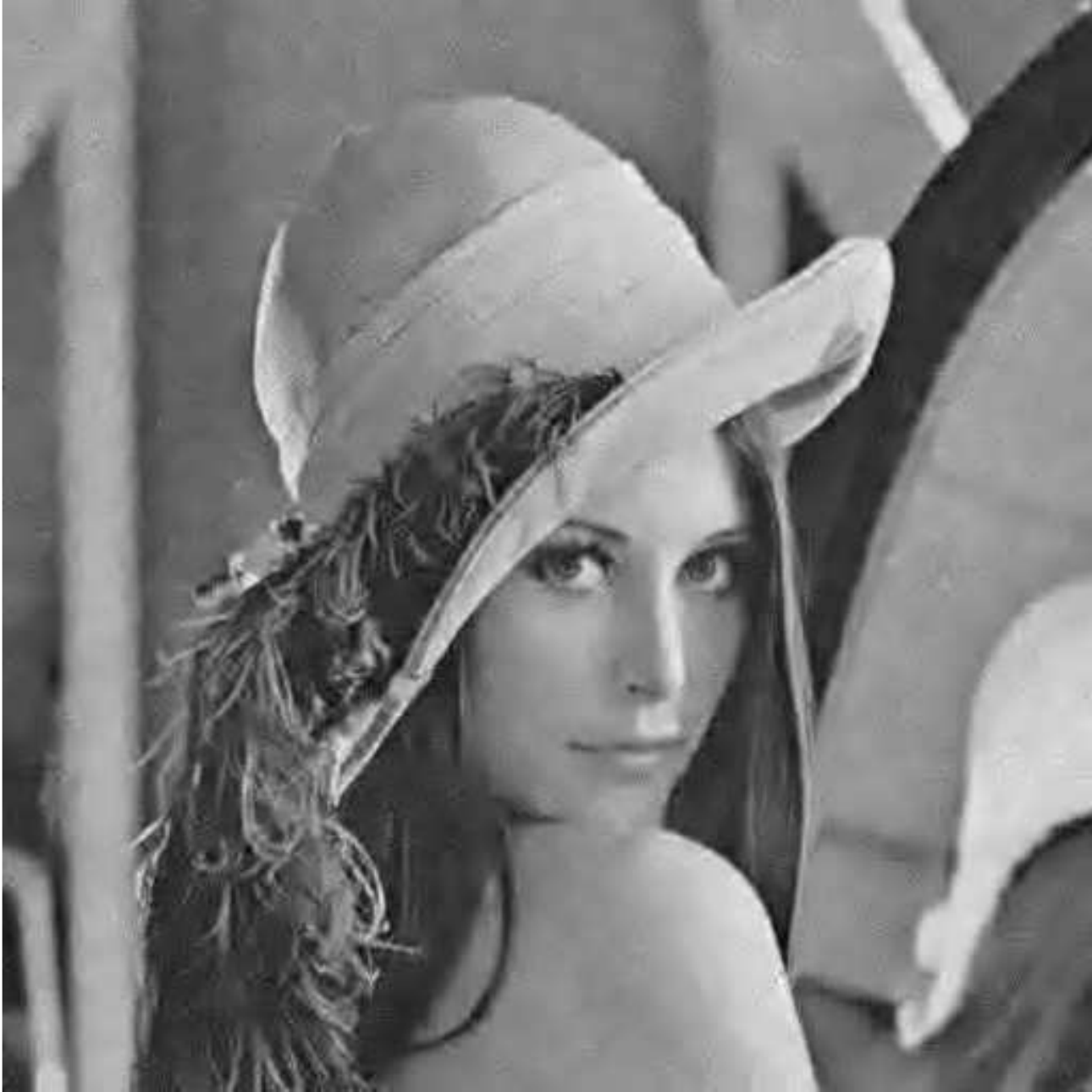}
}
\subfigure[Image denoised using the dictionary trained with Algorithm 1, PSNR $32.22$ dB]{
\includegraphics[width=3.7cm]{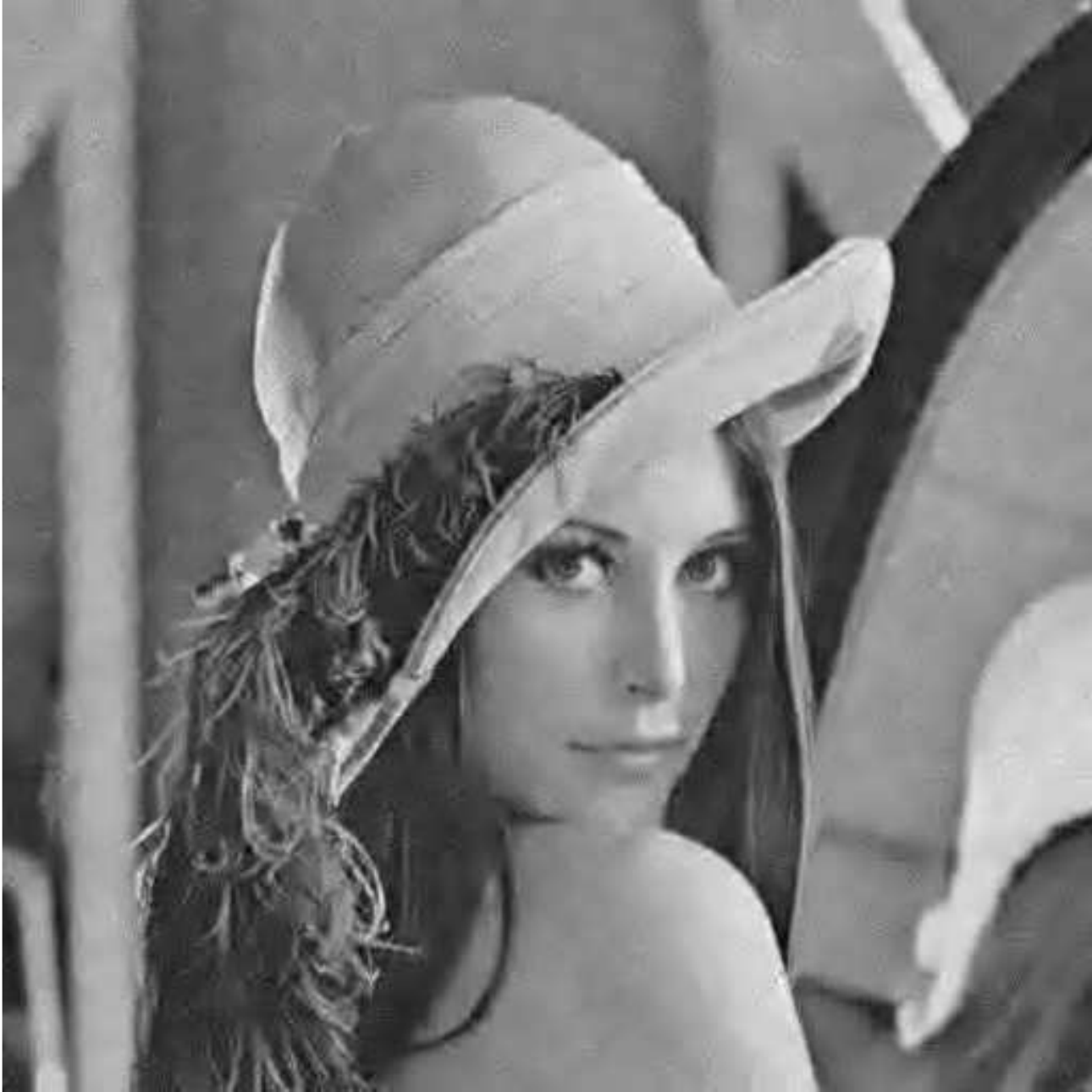}
}
\caption{\small Comparison of denoising performance on an image from the IISc. building database and the standard `Lenna' image.\vspace{-0.6cm}}
\label{test_result_fig}
\end{figure*}


\begin{table}[b]
\centering

\begin{tabular}{|| c || c|| c|| }
\hline
\hline
Performance & Standard &  Split-and-  \\
(on training dataset)& approach &Merge \\
& & algorithm\\
\hline
\hline

Overall MSE (dB) & & \\

\scriptsize $MSE_{train} = \displaystyle\frac{\left\|Y_{train}-\hat{D}\hat{X}\right\|_F}{\left\| Y_{train} \right\|_F}$  & $-21.18$  & $-15.91$   \\

\hline
\hline
Atom recovery accuracy (\%)  & $94.27$  & $85.53$   \\
& &\\
\hline
\hline
Training time (s)& 158.16  & 5.95  \\
\hline
\hline
%
%
\end{tabular}
\caption{\small Performance comparison of the conventional and the proposed parallel learning approach on synthesized signals. The MSE value on the training dataset and atom recovery accuracy are averaged over 20 independent trials.}
\label{table2}
\end{table}

\begin{table}[t]
\centering

\begin{tabular}{||   c        ||   c   ||   c   ||  c || }
\hline
\hline

Input & Lenna  & Boats & House  \\
$\text{PSNR}$&   &    &     \\
\hline
\hline
 $ 28.13 $ &    $35.44 \,\,$   \vline  $\,\, \textcolor{red}{35.40}$    &     $33.56 \,\,$   \vline  $\,\,\textcolor{red}{33.46} $ &      $35.61 \,\,$   \vline  $\,\, \textcolor{red}{35.46}$    \\

\hline
\hline
$ 24.61$ &   $ 33.59\,\,$   \vline  $\,\, \textcolor{red}{33.58}$  &    $31.70 \,\,$   \vline  $\,\,\textcolor{red}{31.62} $ &      $ 33.75\,\,$   \vline  $\,\, \textcolor{red}{33.70}$          \\

\hline
\hline
$ 22.11$ &   $32.21 \,\,$   \vline  $\,\, \textcolor{red}{32.22}$    &   $30.29 \,\,$   \vline  $\,\,\textcolor{red}{30.24} $  &      $ 32.50\,\,$   \vline  $\,\,\textcolor{red}{32.47} $      \\

\hline
\hline
 $20.17$ &    $ 31.08\,\,$   \vline  $\,\,\textcolor{red}{31.10} $   &   $29.21 \,\,$   \vline  $\,\, \textcolor{red}{29.20}$    &      $31.32 \,\,$   \vline  $\,\, \textcolor{red}{32.32}$       \\

\hline
\hline
 $14.15$ &    $27.44 \,\,$   \vline  $\,\, \textcolor{red}{27.50}$   &  $ 25.82\,\,$   \vline  $\,\, \textcolor{red}{25.85}$  &      $27.27 \,\,$   \vline  $\,\, \textcolor{red}{27.28}$       \\

\hline
\hline
\end{tabular}

\caption{\small  (Colour online) Denoising performance on standard images: PSNR values of the noisy input images and denoised output images obtained using the dictionary learned with the conventional approach and the proposed parallel learning approach, averaged over $5$ independent noise realizations. In each cell, the left entry (in black) corresponds to the conventionally trained dictionary and the right entry (in red) corresponds to the dictionary trained using Algorithm 1. }
\label{table1}
\end{table}

\begin{table*}[t]
\centering

\begin{tabular}{||   c        ||   c   ||   c   ||  c   ||   c   ||   c  ||   c  ||   }
\hline
\hline

$\sigma/\text{PSNR}$& Building30  & Building31 & Building32 & Building33 & Building34 & Average  \\
\hline
\hline
 
$10/28.13$ &    $33.28 \,\,\,$   \vline  $\,\,\, \textcolor{red}{33.10}$   &   $35.36 \,\,\,$   \vline  $\,\,\, \textcolor{red}{35.07}$ &    $35.39 \,\,\,$   \vline  $\,\,\,\textcolor{red}{34.95} $ &   $ 34.38\,\,\,$   \vline  $\,\,\,\textcolor{red}{34.13} $  &   $34.10 \,\,\,$   \vline  $\,\,\,\textcolor{red}{33.88} $     &       $34.50 \,\,\,$   \vline  $\,\,\, \textcolor{red}{34.23}$       \\
\hline
\hline
$15/24.61$ &   $30.91 \,\,\,$   \vline  $\,\,\,\textcolor{red}{30.72} $  &    $33.15 \,\,\,$   \vline  $\,\,\,\textcolor{red}{32.91} $  &   $33.22 \,\,\,$   \vline  $\,\,\, \textcolor{red}{32.81}$  &   $32.19 \,\,\,$   \vline  $\,\,\,\textcolor{red}{31.96} $  &    $31.88 \,\,\,$   \vline  $\,\,\,\textcolor{red}{31.66} $  &    $ 32.27\,\,\,$   \vline  $\,\,\, \textcolor{red}{32.01}$           \\

\hline
\hline
$20/22.11$ &   $29.38 \,\,\,$   \vline  $\,\,\,\textcolor{red}{29.19} $    &   $ 31.52\,\,\,$   \vline  $\,\,\, \textcolor{red}{31.37}$  &   $31.75 \,\,\,$   \vline  $\,\,\, \textcolor{red}{31.36}$    &    $30.58 \,\,\,$   \vline  $\,\,\,\textcolor{red}{30.39} $ &   $ 30.36\,\,\,$   \vline  $\,\,\, \textcolor{red}{30.21}$      &    $ 30.72\,\,\,$   \vline  $\,\,\, \textcolor{red}{30.50}$       \\

\hline
\hline
 $25/20.17$ &    $ 28.23\,\,\,$   \vline  $\,\,\,\textcolor{red}{28.09} $   &   $30.30 \,\,\,$   \vline  $\,\,\,\textcolor{red}{30.21} $   &   $30.56 \,\,\,$   \vline  $\,\,\, \textcolor{red}{30.25}$   &   $ 29.36\,\,\,$   \vline  $\,\,\, \textcolor{red}{29.19}$   &    $29.30 \,\,\,$   \vline  $\,\,\, \textcolor{red}{29.19}$    &      $29.55 \,\,\,$   \vline  $\,\,\, \textcolor{red}{29.39}$        \\

\hline
\hline
 $50/14.15$ &    $24.62 \,\,\,$   \vline  $\,\,\, \textcolor{red}{24.58}$   &  $26.45 \,\,\,$   \vline  $\,\,\, \textcolor{red}{26.45}$    &    $ 26.64\,\,\,$   \vline  $\,\,\,\textcolor{red}{26.56}$  &    $25.58 \,\,\, $   \vline  $ \,\,\,\textcolor{red}{25.55} $  &     $26.02 \,\,\, $   \vline  $\,\,\, \textcolor{red}{25.99}$   &     $25.86 \,\,\,$   \vline  $\,\,\, \textcolor{red}{25.83}$        \\

\hline
\hline
\end{tabular}

\caption{\small (Colour online) Denoising performance on IISc. building images: PSNR values (in dB) of the noisy input images and denoised output images obtained using the dictionary learned with the conventional approach and the split-and-merge approach, averaged over $5$ independent noise realizations. In each cell, the left entry (in black) corresponds to the conventionally trained dictionary and the right entry (in red) corresponds to the dictionary trained using the proposed algorithm. \vspace{-0.4cm}}
\label{table1}
\end{table*}


\subsection{Synthesized signal experiment}
\label{ssec:subhead}
We create a matrix $D$ of size $30 \times 60$ with i.i.d. samples of the Gaussian distribution with zero mean and unity variance (denoted by $\mathcal{N}(0,1)$), and normalize the columns so that they have unit length. Subsequently, we produce $N=4\times10^4$ training examples, each by taking random combinations of $s=6$ atoms in $D$, with coefficients drawn from the $\mathcal{N}(0,1)$ distribution. For training using the standard approach, we initialize the dictionary by taking the first $60$ training vectors as the dictionary atoms. To train the dictionary using the split-and-merge algorithm, we partition the entire dataset into $L=40$ smaller datasets, each having $n=\frac{N}{L}=10^3$ training vectors. Over each of the smaller datasets, we train a dictionary of size $30 \times 50$ for a sparse representation with sparsity $s_1=3$. Note that $K_1L=2\times 10^3$, and $\frac{N}{L}=10^3$ have the same order of magnitude. The dictionaries are merged into a single dictionary of size $30 \times 60$ using the approach described in step 3 of Algorithm 1, where we chose $s_2=\frac{s}{s_1}=2$. Since the ground truth is known for the synthetic experiment, we measure the closeness of the recovered dictionary with the actual one in the following manner: we declare that an atom $d_i$ has been recovered from the true dictionary $D$, if there exists an atom $\hat{d}_j$ in the estimated dictionary $\hat{D}$ such that $\left|  d_i^T \hat{d}_j \right| \geq 0.98$. 

\noindent The performance of the proposed algorithm on the synthesized training dataset is shown is Table 1. The values of the mean squared error (MSE) on the training dataset and the accuracy of atom recovery are averaged over $20$ independent trials. As shown by our experimental results, the split-and-merge technique results in an increment of  MSE by approximately $6$ dB and a reduction in atom recovery accuracy by $8$\%, but the training time reduces drastically, almost by a factor of $26$. The deterioration of performance in terms of MSE and atom recovery accuracy is acceptable for most practical purposes, and the remarkable reduction in training time makes it suitable for many big data applications. 
\subsection{Image denoising}
\label{ssec:subhead}
For the purpose of comparing the proposed parallel dictionary learning algorithm with the usual learning strategy, we consider the task of denoising images corrupted by additive white Gaussian noise with variance $\sigma^2$. We train the dictionary on a database of clean image patches of size $8 \times 8$ and use the same to estimate the clean image from the noisy input.~We create two databases of clean image patches: first one with clean patches from the IISc.~building images (the database is available from the authors upon request), and the second one with clean patches from images that are frequently used in image processing applications. The dictionaries are tested on images which do not belong to the training database. We report the denoising performance of the dictionary trained using Algorithm 1 on the building images as well as on the standard images. The details of the training and denoising processes are given in the following two subsections.
\subsubsection{Training}
To train the dictionary, we use a database consisting of clean images of the IISc. buildings and standard images, and extract $10^5$ patches randomly from them, each of size $8 \times 8$. Both standard and the split-and-merge algorithms are initialized with an overcomplete DCT dictionary of size $64 \times 256$ and the iterations are repeated $100$ times. While deploying the split-and-merge algorithm for dictionary learning, the database of $10^5$ patches is divided into $20$ smaller datasets, each containing $5000$ patches, and over each of them, we train a dictionary of size $64 \times 128$. The locally trained dictionaries are merged together into a global dictionary of size $64 \times 256$. The time taken to train the dictionary using the conventional approach and the parallel learning approach is $1.25\times 10^3$ seconds and $33.59$ seconds, respectively. 
\vspace{-0.3cm}
\subsubsection{Denoising}
In the denoising phase, we extract all noisy patches (with overlap of $1$ pixel in both horizontal and vertical directions) of size $8 \times 8$ from the given image and solve the following sparse coding problem using the OMP algorithm: 
$\hat{x} = \arg \underset{x}{\min} \left\| x \right\|_0 \text{\,\,such that\,\,} \left\| y-\hat{D}x \right\|_2 \leq \epsilon$, where $y$ denotes the noisy image patch, $\hat{D}$ is the dictionary trained on the database of clean images, and $\hat{x}$ denotes the estimate of the clean image patch. We experimentally observed that $\epsilon=8.5\sigma$ is optimum, where $\sigma^2$ is the variance of the additive Gaussian noise corrupting the image. After obtaining the estimates of the clean image patches corresponding to all noisy patches, we take the average of the overlapping estimated patches to obtain the denoised output image. The results of the denoisng experiment are reported in Figure 1 and Tables 2 and 3. We show a comparison of the PSNR values of the denoised images averaged over $5$ independent noise realizations, using the dictionaries learned with the standard and the proposed approaches. We observe that the dictionary trained using the split-and-merge algorithm is on par with its conventionally (using the whole data at a time) trained counterpart in terms of PSNR of the denoised output, but results in a reduction in training time approximately by a factor of $37$.
\vspace{-0.2cm}
\section{CONCLUSION}
\vspace{-0.3cm}
We have proposed a parallel dictionary learning algorithm for sparse representation of a set of training vectors. The basic philosophy behind our algorithm is to partition the big dataset into smaller ones, learn a dictionary over each of the smaller datasets, and finally, combine them into a single dictionary that represents the whole dataset using a coefficient matrix having sparse columns. The parallel learning approach performs on par with the conventional learning strategy, as indicated by the experimental results on synthesized signals as well as by image denoising experiments. The PSNR values of the denoised images using the proposed algorithm fall short by only $0.1-0.2$ dB on an average, as compared with the PSNR values obtained using the conventionally trained dictionary. The key advantage with the parallel learning algorithm is that it involves less computational complexity (c.f. Appendix B) compared with the conventional approach, thereby facilitating faster learning adapted to data.  
\label{sec_conclusion}
\subsubsection*{Appendix A: Proof of Proposition 1}
Since $y \in \mathcal{S}\left( \tilde{D},x_1,\epsilon_1,s_1 \right)$, we have from Definition 1 that $y = \tilde{D} x_1+e_1$, with $\left\| x_1 \right\|_0 \leq s_1$ and $\left\| e_1 \right\|_2 \leq \epsilon_1$. Using the fact that each column $ \tilde{d}_j$ of $ \tilde{D}$ belongs to $\mathcal{S}\left( D,z_j,\frac{\epsilon_2}{\left\| x_1 \right\|_2},s_2 \right)$, we write $ \tilde{D} = DZ+E$, where each column of $Z$ satisfies $\left\| z_j \right\|_0 \leq s_2$, and the columns of $E$ satisfy $\left\| e_j \right\|_2 \leq \frac{\epsilon_2}{\left\| x_1 \right\|_2}$. Therefore, we get
\begin{eqnarray*}
y =  \tilde{D} x_1+e_1 
&=& \left(DZ + E\right) x_1 + e_1\\
&=& Dr + Ex_1 + e_1,
\end{eqnarray*}
where $r =Zx_1= \sum_{j=1}^{K}x_{1j}z_j$ has atmost a sparsity of $s=s_1s_2$, with $x_{1j}$ being the $j^{\text{th}}$ entry of $x_1$. Observe that the sparsity of $r$ is at most $s_1s_2$, because of the fact that it is a linear combination of $s_1$ number of $s_2$-sparse signals, and 
\begin{eqnarray*}
 \left\| y-Dr \right\|_2 &=& \left\| Ex_1 + e_1\right\|_2 \\
&\leq& \left\| E \right\|_F \left\| x_1 \right\|_2 +  \left\| e_1 \right\|_2\\
&\leq& K  \frac{\epsilon_2}{\left\| x_1 \right\|_2}\left\| x_1 \right\|_2 + \epsilon_1=K\epsilon_2+ \epsilon_1\\
\end{eqnarray*}
Hence, we have that $y \in \mathcal{S}\left( D,Zx_1,K\epsilon_2+\epsilon_1,s_1 s_2 \right)$.
\subsubsection*{Appendix B: Computational Complexity of Algorithm 1}

{\bf Sparse coding:} For a dataset of size $n$, the sparse coding step using OMP requires $\mathcal{O}\left( smKn \right)$ computations \cite{elad2}.

\noindent {\bf Dictionary update:} For a dictionary of size $m \times K$, in this step one computes the SVD of an $m \times n$ matrix, which requires $\mathcal{O}\left(m^2n+n^3 \right)$ operations. Therefore, the total computation time required for each iteration is given by $T=cn\left( Ksm+m^2+n^2\right)$, for some constant $c>0$. Let $T_1$ and $T_2$ be the computation times required in each iteration of the standard learning approach and the smaller subproblems, respectively. Then, we have that 

\begin{eqnarray*}
T_1 &=& cN \left( Ksm+m^2+N^2  \right), \text{\,\,and}\\
T_2 &=& c\frac{N}{L} \left( K_1s_1m+m^2+\frac{N^2}{L^2}  \right).
\end{eqnarray*}
Therefore, the total time taken for each iteration of the subproblems and the merging step is given by
 \scriptsize
 \begin{eqnarray*} 
T_{\text{total}} &=& cN \left( K_1s_1m+m^2+\frac{N^2}{L^2}  \right) + cK_1L \left( Ks_2m +m^2+K_1^2 L^2 \right)\\
&\stackrel{(a)}{\leq}& cN \left( K_1s_1m+m^2+\frac{N^2}{L^2}  \right) + c\frac{N}{L} \left( Ks_2m +m^2+\frac{N^2}{L^2}\right)\\
&\ll& T_1,
\end{eqnarray*}
\small
where $(a)$ follows from the assumption that $K_1L \approx \frac{N}{L}$. 

\bibliographystyle{IEEEbib}
\bibliography{strings,refs}

\vfill
\pagebreak

\bibliographystyle{IEEEbib}
\bibliography{strings,refs}

\end{document}